\documentclass[final]{cvpr}

\usepackage{soul}
\usepackage{setspace}
\usepackage{url}
\usepackage{booktabs}
\usepackage{algorithm}
\usepackage{algorithmic}
\usepackage{multirow}
\usepackage{subfigure}
\usepackage{adjustbox}

\usepackage{times}
\usepackage{epsfig}
\usepackage{graphicx}
\usepackage{amsmath}
\usepackage{amssymb}

\usepackage[pagebackref=true,breaklinks=true,letterpaper=true,colorlinks,bookmarks=false]{hyperref}

\def\cvprPaperID{****} 
\def\confYear{CVPR 2021}

\begin{document}

\title{EfficientPose: Efficient Human Pose Estimation with Neural Architecture Search}

\author{Wenqiang Zhang$^{1}$\footnotemark[1] , Jiemin Fang$^{2,1}$\footnotemark[1] , Xinggang Wang$^1$\footnotemark[2] , Wenyu Liu$^1$\\
$^1$School of EIC, Huazhong University of Science \& Technology\\
$^2$Institute of Artificial Intelligence, Huazhong University of Science \& Technology\\
\texttt{\small\{wq\_zhang, jaminfong, xgwang, liuwy\}@hust.edu.cn}\\
}

\maketitle

\begin{abstract}
   Human pose estimation from image and video is a vital task in many multimedia applications. Previous methods achieve great performance but rarely take efficiency into consideration, which makes it difficult to implement the networks on resource-constrained devices. Nowadays real-time multimedia applications call for more efficient models for better interactions. Moreover, most deep neural networks for pose estimation directly reuse the networks designed for image classification as the backbone, which are not yet optimized for the pose estimation task. In this paper, we propose an efficient framework targeted at human pose estimation including two parts, the efficient backbone and the efficient head. By implementing the differentiable neural architecture search method, we customize the backbone network design for pose estimation and reduce the computation cost with negligible accuracy degradation. For the efficient head, we slim the transposed convolutions and propose a spatial information correction module to promote the performance of the final prediction. In experiments, we evaluate our networks on the MPII and COCO datasets. Our smallest model has only 0.65 GFLOPs with 88.1\% PCKh@0.5 on MPII and our large model has only 2 GFLOPs while its accuracy is competitive with the state-of-the-art large model, i.e., HRNet with 9.5 GFLOPs.
\end{abstract}
\renewcommand{\thefootnote}{\fnsymbol{footnote}}
\footnotetext[1]{Equal contribution.}
\footnotetext[2]{Corresponding author.}
\renewcommand{\thefootnote}{\arabic{footnote}}


\section{Introduction}

\begin{figure}
\centering
\includegraphics[height=6.15cm,width=8.2cm]{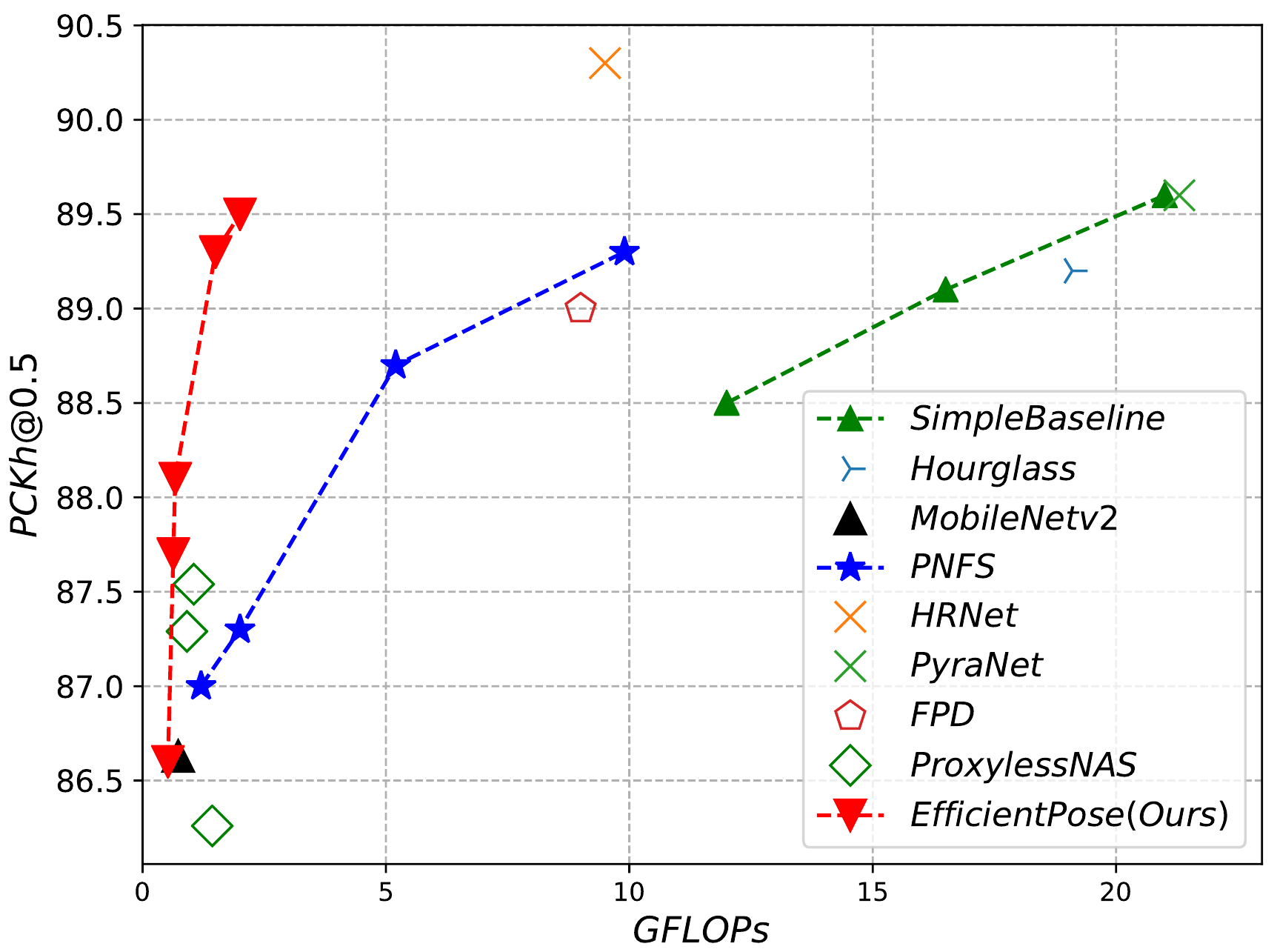}
\caption{FLOPs comparisons of different networks on the MPII validation dataset.}
\label{fig: comp}
\vspace{-6.5pt}
\end{figure}

Human pose estimation has a wide range of multimedia applications in the real world, e.g., virtual reality, game interaction and assisted living. Human pose estimation targets at predicting the locations of human keypoints or parts in images. Traditional methods~\cite{yang2011articulated,pish2013poselet} are based on probabilistic graphical model or pictorial structures. Deep convolution neural networks boost the development of this field by directly regressing keypoints positions~\cite{toshev2014deeppose} or predicting the heatmap locations~\cite{newell2016stacked}, solving this problem in an end-to-end manner. Latter top-down methods~\cite{xiao2018simple,sun2019deep} achieve high accuracies on both MPII~\cite{andriluka20142d} and COCO~\cite{COCO} benchmarks. However, the backbone networks in most previous methods~\cite{xiao2018simple,Chen2018CPN,li2019rethinking} often directly reuse the networks designed for image classification, e.g., ResNet~\cite{he2016deep}, VGG~\cite{simonyan2014very}, which result in sub-optimality for human pose estimation tasks. Moreover, existing human pose estimation methods over-pursue the accuracy but ignore the efficiency of the model, making it difficult to deploy the model on the resource-constrained computing devices commonly used in real-life scenarios. Recent multimedia applications require more efficient human pose estimation which can bring better interaction experience for users. However, current pose estimation algorithms cannot meet this requirement.

In recent years, the emergence of neural architecture search (NAS) methods greatly accelerates the development of neural network design. Some pioneering works~\cite{zoph2018learning,real2019regularized} take huge search cost to obtain an architecture with high accuracies on image classification tasks. Latter one-shot and differentiable NAS methods \cite{Understanding,liu2018darts} greatly decrease the search cost while keeping the high performance of the searched architectures. NAS is also applied to cut down the FLOPs or latency of the architecture while the high accuracy is guaranteed~\cite{cai2018proxylessnas,fbnet}. Furthermore, some methods aim at directly searching on the target tasks for customized architecture design and performance promotion, e.g., semantic segmentation~\cite{liu2019auto,zhang2019customizable} and object detection~\cite{ghiasi2019fpn,fang2020fast,fang2020fna++}. However, searching for the backbone networks for segmentation, detection and pose estimation is very computational expensive. 
In previous pose estimation methods, the customized and automatic backbone design is rarely explored.
For more optimized networks, searching for the backbone is more important as the backbone takes up a large part of the whole network and plays the role of feature extraction.

\begin{figure*}[h]
\centering
    \includegraphics[width={0.97\textwidth}]{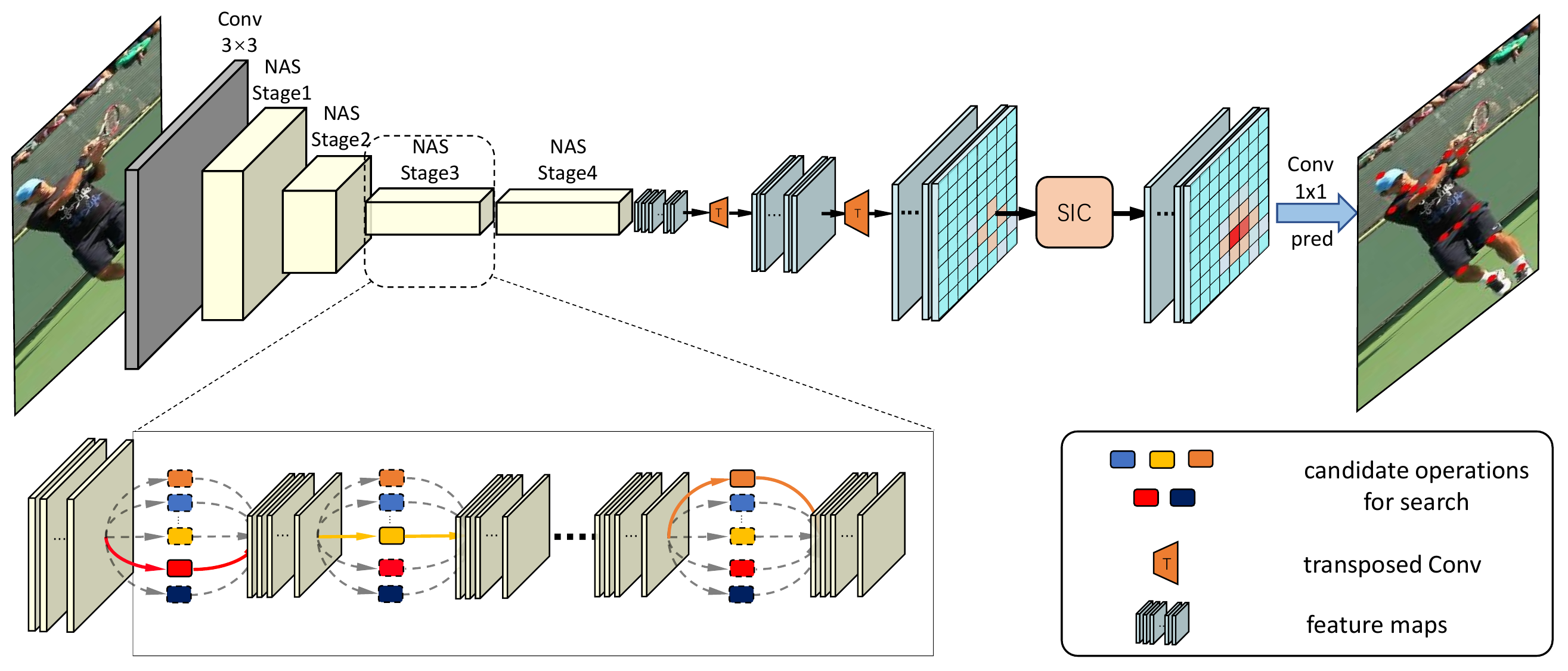}
    \caption{The framework of EfficientPose. For the backbone part, we use the NAS method to obtain more efficient and precise networks. The desired network is finally derived from the super network. For the head part, we slim the transposed convolution and use a SIC module to correct the spatial information in the feature maps after the transposed convolutions. The input ``Conv$3\times3$" denotes a plain $3\times3$ convolution followed by a $3\times3$ separable depthwise convolution~\cite{howard2017mobilenets}.}
    \label{fig: framework}
\end{figure*}

To directly search for the backbone for pose estimation, we propose to design an efficient human pose estimation network searching framework, namely EfficientPose. As shown in Fig.~\ref{fig: framework}, our efficient network includes two main parts, efficient backbone and efficient head. To tackle the backbone performance bottleneck in terms of both accuracy and efficiency, the differentiable NAS method~\cite{liu2018darts} is designed for lightening the computation cost of the backbone network and adapting the backbone architecture to pose estimation tasks. Moreover, we design an efficient pose estimation head which enables not only fast inference but also fast architecture search. In the head network, the transposed convolutions are made slimmer according to the width of the backbone. A \emph{spatial information correction} (SIC) module is proposed to promote the quality of the feature maps used for generating the heatmaps. Overall, we promote the efficiency of the whole framework for pose estimation. The architectures generated by our framework produce high performance with low FLOPs and latencies, as shown in Fig.~\ref{fig: comp}. 

Our contributions can be summarized as follows:
\begin{itemize}
    \item We propose an efficient pose estimation network searching framework, which is the first method that performs backbone search for pose estimation. The differentiable NAS method adjusts and lightens the backbone network automatically. The head network is redesigned by the proposed slim transposed convolutions and spatial information correction module to get a better trade-off between computation cost and accuracy.
    \item We obtain an extremely efficient pose estimation network that costs only 0.65 GFLOPs with high performance (88.1\% PCKh@0.5). Our large model has only 2 GFLOPs while its accuracy is competitive with the state-of-the-art large model, i.e., HRNet~\cite{sun2019deep} with 9.5 GFLOPs. Besides, the good generalization ability of the proposed EfficientPose networks has been confirmed on the COCO dataset.
\end{itemize}

\section{Related Work}
\paragraph{Human Pose Estimation}
Previous methods of human pose estimation have achieved great success. DeepPose~\cite{toshev2014deeppose} first applies deep learning to pose estimation tasks, regarding pose estimation task as regression problems. The convolutional neural network finally outputs the keypoints positions directly. Most of work in recent years predict the location of heat maps, which allows the network to save more spatial information. Hourglass~\cite{newell2016stacked} proposes to stack repeating U-shape blocks with skip connections to promote overall performance. PyraNet~\cite{yang2017pyramid} uses a pyramid residual module to obtain multi-scale information. CPN~\cite{Chen2018CPN} adopts a GlobalNet to roughly localize keypoints and a RefineNet to explicitly handle hard keypoints. SimpleBaseline~\cite{xiao2018simple} aims at constructing a simplified network and applies transposed convolutions to get high-resolution heatmaps. HRNet~\cite{sun2019deep} achieves state-of-the-art results on COCO~\cite{COCO} by maintaining high-resolution representations, while the multi-branch framework produces high computation cost and is not friendly to inference on embedded devices. RSN~\cite{cai2020learning} continues to push forward the results of this task, which uses attention mechanism and features aggregation with same level to get precise keypoint localization.

Though previous methods have achieved high accuracies in human pose estimation tasks, most of them bear large computation cost and high latency. The efficiency of the network is rarely considered, which makes it difficult to be implemented in real-word scenarios. Bulat et al.~\cite{bulat2017binarized} promote the performance of the binarized model to get a better trade-off between model size and accuracy. DU-Net~\cite{tang2018quantized} applies the quantification method to reduce the model size. FPD~\cite{Zhang_2019_CVPR} uses a strong teacher network to supervise the small student networks for performance promotion. We aim at constructing a lightweight framework for pose estimation which achieves the high accuracy with a low computation cost. 

\paragraph{Neural Architecture Search}
Recent NAS methods greatly promote the performance of neural networks. Early NAS works use reinforcement learning~\cite{zoph2018learning,tan2019efficientnet} or the evolutionary algorithm~\cite{real2019regularized,fang2019eat} to search for the architectures and achieve superior results than networks designed manually. Latter one-shot~\cite{Understanding,brock2017smash} or differentiable~\cite{liu2018darts,cai2018proxylessnas,fbnet,dong2019searching,xu2019pc} NAS methods search for high-performance architectures with low computation cost. To further cut the search cost, some methods search for a cell structure and then stack it to build the final architecture~\cite{liu2018darts,xu2019pc}. NAS is also applied on promoting the model efficiency~\cite{MnasNet,tan2019efficientnet,cai2018proxylessnas,fbnet} by optimizing the FLOPs, latency etc. Meanwhile, NAS methods have been implemented on different tasks, e.g. semantic segmentation~\cite{liu2019auto,zhang2019customizable} and object detection~\cite{ghiasi2019fpn,fang2020fast,fang2020fna++}. In this paper, we implement differentiable NAS on the backbone network design targeted at human pose estimation. Our networks with NAS adjusted achieve a better trade-off between accuracy and efficiency compared to other state-of-the-art methods~\cite{xiao2018simple,sun2019deep}.


\section{Method}
In this section, we first introduce the method of differentiable neural architecture search, which is used for designing the backbone network automatically targeted at pose estimation. Secondly, we explain how the search space for the backbone network is designed and how we optimize both accuracy and latency. Finally, we illustrate the designing principles of our redesigned lightweight head. The whole framework is shown in Fig.~\ref{fig: framework}.

\subsection{Differentiable Neural Architecture Search}

We use differentiable NAS \cite{liu2018darts,cai2018proxylessnas,fbnet} to customize the backbone network design on the pose estimation task. We formulate the NAS problem as a nested optimization problem
\begin{equation}
\min \limits_{a\in \mathcal{S}}\min \limits_{w_{a}} \mathcal{L}\left(a, w_{a}\right)\text{,}
\end{equation}
where $\mathcal{S}$ represents the search space and $w_a$ denotes the operation weights of architecture $a$. We search for the architecture $a$ by minimizing the loss $\mathcal{L}(a, w_a)$. 

In the differentiable NAS method, the search space $\mathcal{S}$ is relaxed into a continuous representation. Specifically, in every layer of the architecture, we compute the probability of each candidate operation as
\begin{equation}
p_{o}^{\ell} = \frac{exp(\alpha_{o}^{\ell})}{\sum_{o^{\prime} \in \mathcal{O}}exp({\alpha}_{o^{\prime}}^{\ell})}\text{,}
\end{equation}
where $\mathcal{O}$ denotes the set of candidate operations and $\alpha_o^{\ell}$ denotes the architecture parameter of operation $o$ in layer $\ell$. The output of each layer is computed as a weighted sum of output tensors from candidate operations
\begin{equation}
    x_{\ell+1} = \sum_{o \in \mathcal{O}} p_{o}^{\ell} \cdot o(x_{\ell})\text{,}
\end{equation}
where $x_{\ell}$ denotes the input tensor of layer $\ell$. 

During the search process, the operation weights and architecture parameters are optimized alternately by stochastic gradient descent to minimize the loss $\mathcal{L}$. The final searched architecture is derived based on the distribution of architecture parameters.

\subsection{Efficient Backbone}

Most previous methods~\cite{wei2016cpm,Chen2018CPN,xiao2018simple,li2019rethinking} use image classification networks as the backbone for pose estimation, e.g., ResNet~\cite{he2016deep}, MobileNetV2~\cite{sandler2018mobilenetv2}. To narrow the gap between classification and pose estimation tasks, we redesign the backbone network with the NAS method. We first give the details about the search space. Then we describe the method of optimizing both accuracy and latency of the model.

\begin{table}[thbp]
\centering
\caption{Search space of the backbone. The first ``Conv3$\times$3" denotes the plain convolution with a 3$\times$3 kernel size. ``SepDepth$3\times3$" denotes a separable depthwise convolution~\cite{howard2017mobilenets} with kernel size $3\times3$. ``TransConv" denotes the transposed convolution. s2 denotes the stride of 2. EfficientPose-A is searched under the small setting of the search space, while EfficientPose-B and -C are searched under the large setting.}
\resizebox{\columnwidth}{!}{
    \begin{tabular}{l |c |c |c}
    \hline
    \multirow{2}*{\textbf{Stage}} & \multirow{2}*{\textbf{Output Size}} & \multicolumn{2}{c}{\textbf{Layer}} \\
    \cline{3-4}
    && Small & Large \\
    \hline
    Input          & $256\times256$        & \multicolumn{2}{c}{-} \\
    \hline
    Conv3$\times$3 & $128\times128$ &\multicolumn{2}{c}{[32, s2]} \\
    \hline
    SepDepth3$\times$3 & $128\times128$ & [16, s1] & [24, s1] \\
    \hline
    \multirow{2}*{NAS Stage1} & \multirow{2}*{$64\times64$} & [24, s2] & [32, s2] \\
    & & [24, s1]$\times$3 & [32, s1]$\times$5 \\
    \hline
    \multirow{2}*{NAS Stage2}   & \multirow{2}*{$32\times32$} & [32, s2] & [64, s2] \\
    & & [32, s1]$\times$5    & [64, s1]$\times$7 \\
    \hline
    \multirow{2}*{NAS Stage3}   & \multirow{2}*{$16\times16$} & [64, s2] & [96, s2] \\
    & & [64, s1]$\times$9   & [96, s1]$\times$9 \\
    \hline
    NAS Stage4   & $16\times16$ & [96, s1]$\times$8 & [160, s2]$\times$10 \\
    \hline
    TransConv & $32\times32$ &\multicolumn{2}{c}{[64, s2]} \\
    \hline
    TransConv & $64\times64$ &\multicolumn{2}{c}{[32, s2]} \\
    \hline
    \end{tabular}
    }
\label{tab: Search_space}
\end{table}

\paragraph{Search Space}
We construct our search space based on the MobileNetV2~\cite{sandler2018mobilenetv2} network, which holds the great performance with low computation cost and is commonly used to design the search space in NAS methods~\cite{cai2018proxylessnas,fbnet}. We stack inverted residual blocks (i.e., MBConvs) to build the backbone network and allow MBConvs to have various settings. Specifically, the kernel sizes include \{3, 5, 7\} and expansion ratios include \{3, 6\}. The skip connection is used in the candidate operations to search the depth of the network. The dropping-path training strategy~\cite{Understanding,cai2018proxylessnas,fang2020densely} is performed to decrease the coupling effect between different sub-architectures in the search space. 

Different from previous methods~\cite{xiao2018simple,sun2019deep}, we only perform 4 down-sampling operations in the backbone network (5 in most other methods). We consider that in the lightweight pose estimation network, 5$\times$ down-sampled feature maps contribute little to the final prediction and may lose much information. The higher-resolution (4$\times$) feature maps are used for up-sampling. We give the details of the backbone network in Tab.~\ref{tab: Search_space}.

\paragraph{Cost Optimization}
The cost optimization of a neural network is of critical importance and can be measured by different benchmarks, e.g., FLOPs, latency. To obtain the high-performance network with low cost, we integrate the cost optimization into the search formulation. As most differentiable NAS methods~\cite{cai2018proxylessnas,fbnet,fang2020densely} do, we build a lookup table to predict the cost of the architecture during search. The cost of one layer is computed as
\begin{equation}
    cost_{\ell} = \sum_{o \in \mathcal{O}}p_{o}^{\ell}{\cdot}cost_{o}^{\ell}\text{,}
\end{equation}
where $cost_{o}^{\ell}$ is the cost of operation $o$ in layer $\ell$ and $p_{o}^{\ell}$ denotes the corresponding probability computed by architecture parameters. The latency of the whole network can be computed as
\begin{equation}
    cost = \sum_{i} cost_i\text{.}
\end{equation}
We add the cost regularization to the loss function for multi-objective optimization. The loss function during the search is defined as
\begin{equation}
\begin{aligned}
    \mathcal{L}\left(a, w_{a}\right) =&\  \mathcal{L}_{MSE} + \lambda \cdot \log_\tau cost \\
    \mathcal{L}_{MSE} =&\  \frac{1}{K} \cdot \sum_{k=1}^{K}{\left\|m_k - \hat{m_k}\right\|_2^2}\text{,}
\end{aligned}
\label{eq: loss}
\end{equation}
where $\hat{m_k}$ is the predicted heatmap of the $k$-th joint while $m_k$ is the ground truth, $\lambda$ and $\tau$ are hyperparameters to balance MSE loss and latency regularization.

\subsection{Efficient Head}
The head of the pose estimation network aims to generate high-resolution heatmaps. To obtain a more efficient network, we redesign the head of the network. We first propose the slim transposed convolutions which produce high-quality heatmaps with computation cost greatly decreased. Then we propose a \emph{spatial information correction} (SIC) module which makes the spatial information of the high-resolution representation more reliable. The SIC module promotes the prediction performance with a negligible computation cost.

\paragraph{Slim Transposed Convolutions}
Different from previous methods~\cite{xiao2018simple,sun2019deep}, the backbone of our network outputs the feature maps with a small number of channels. Accordingly, we cut down the width of the transposed convolutions. Latter experiments show the effectiveness of our slim transposed convolutions. Moreover, we explore more efficient upsampling convolutions in experiments, e.g., we use the separable depthwise convolution~\cite{howard2017mobilenets} to perform upsampling, and achieve great performance as well.

\paragraph{Spatial Information Correction}
As shown in Fig.~\ref{fig: vis_feature}, after the transposed convolutions, the feature maps present a checkerboard pattern of artifacts~\cite{odena2016deconvolution}, which is caused by the uneven overlap of the transposed convolutions. To promote the quality of feature maps for heatmap generating, we add a spatial information correction (SIC) module, a $3\times3$ depthwise convolution in our implementation, after the transposed convolutions. With the SIC module, the checkerboard artifact pattern is almost eliminated which remarkably promotes the performance of the prediction with negligible computation cost. We perform more studies of the module in experiments.

\section{Experiments}
In this section, we first perform the experiments on the MPII  \cite{andriluka20142d} dataset. We describe the implementation details and compare the EfficientPose networks with other state-of-the-art (SOTA) methods and different backbone networks. Then the generalization ability of EfficientPose networks is demonstrated on the COCO~\cite{COCO} dataset. We further specialize EfficientPose network design on different model cost benchmarks. Finally, ablation studies are carried out to show the effectiveness and efficiency of different modules in our framework.

\subsection{Experiments on MPII}
\paragraph{Implementation Details}
The MPII~\cite{andriluka20142d} dataset contains approximately 25K images with about 40K people. Following the standard training settings~\cite{yang2017pyramid,xiao2018simple,sun2019deep}, all input images are cropped to 256$\times$256 for fair comparisons. For the backbone architecture search, we randomly split the training data into two parts, 80\% for operation weight training and 20\% as the validation set to update architecture parameters. The original validation set is never used in the search process. We adopt the same data augmentation strategy as SimpleBaseline~\cite{xiao2018simple}.

Before the search process, we build a lookup table for the latency of each operation. The latency is measured on a single GeForce RTX 2080Ti GPU with a batch size of 32. For the backbone architecture search, we first train the operation weights for 60 epochs which takes 6.5 hours on 2 2080Ti GPUs. Then we start the joint optimization by alternately train operation weights and architecture parameters in each epoch. For operation weight updating, we use the SGD optimizer with 0.9 momentum and 1e-4 weight decay. The initial learning rate is set as 0.05 and is gradually decayed to 0 following the cosine annealing schedule. For architecture parameters optimizing, we use the Adam optimizer with a fixed learning rate of 3e-4. The batch size is set as 32. The joint optimization process takes 150 epochs, 19 hours on 2 2080Ti GPUs. The whole search process only takes 25.5 hours on 2 GPUs, 51 GPU hours in total.

We train the derived network for 200 epochs using the Adam optimizer with the initial learning rate of 1e-3 and a batch size of 32. The learning rate is decayed by 0.1 at 150 epoch and 170 epoch respectively. The fast neural network adaptation method (FNA)~\cite{fang2020fast,fang2020fna++} is implemented for the initialization of both the super network in the search process and the derived network for efficient training.

\begin{table}[t!]
    \centering
    \caption{Comparisons with SOTA methods on the MPII validation set.}
    \vspace{3pt}
    \resizebox{\columnwidth}{!}{
        \begin{tabular}{lrrr}
        \toprule
        Method  & Params & GFLOPs & PCKh@0.5\\
        \midrule
        CPMs~\cite{wei2016cpm}       & 31.0M     & 175.0      & 88.0 \\
        DLCM~\cite{tang2018deeply}    & 15.5M    & 33.6    & 87.5 \\
        SimpleBaseline-R50~\cite{xiao2018simple}      & 34.0M   & 12.0   & 88.5 \\
        PNFS~\cite{yang2019pose}    & - & 2.0  & 87.3 \\
        EfficientPose-A   & 1.3M     & 0.7  & 88.1  \\
        \midrule
        Hourglass~\cite{newell2016stacked}  & 25.1M & 19.1   & 89.2\\
        SimpleBaseline-R101~\cite{xiao2018simple}     & 52.0M   & 16.5   & 89.1\\
        FPD~\cite{Zhang_2019_CVPR}      & 3.0M    & 9.0      & 89.0\\
        PNFS~\cite{yang2019pose}   & - & 9.9 & 89.3\\
        EfficientPose-B  & 3.3M     & 1.5    & 89.3\\
        \midrule
        PyraNet~\cite{yang2017pyramid}    & 28.1M & 21.3   & 89.6\\
        DU-Net~\cite{tang2018quantized}    & 7.9M    & -    & 89.5\\
        DU-Net~\cite{tang2018quantized}    & 15.9M  & -   & 89.9\\
        SimpleBaseline-R152~\cite{xiao2018simple}  & 68.6M & 21.0   & 89.6\\
        HRNet-W32~\cite{sun2019deep}  & 28.5M & 9.5    & 90.3\\
        EfficientPose-C  & 5.0M     & 2.0      & 89.5 \\
        \bottomrule
        \end{tabular}
    }
    \label{tab: mpii results}
\end{table}

\begin{figure}[thbp]
\centering
\includegraphics[scale=0.46]{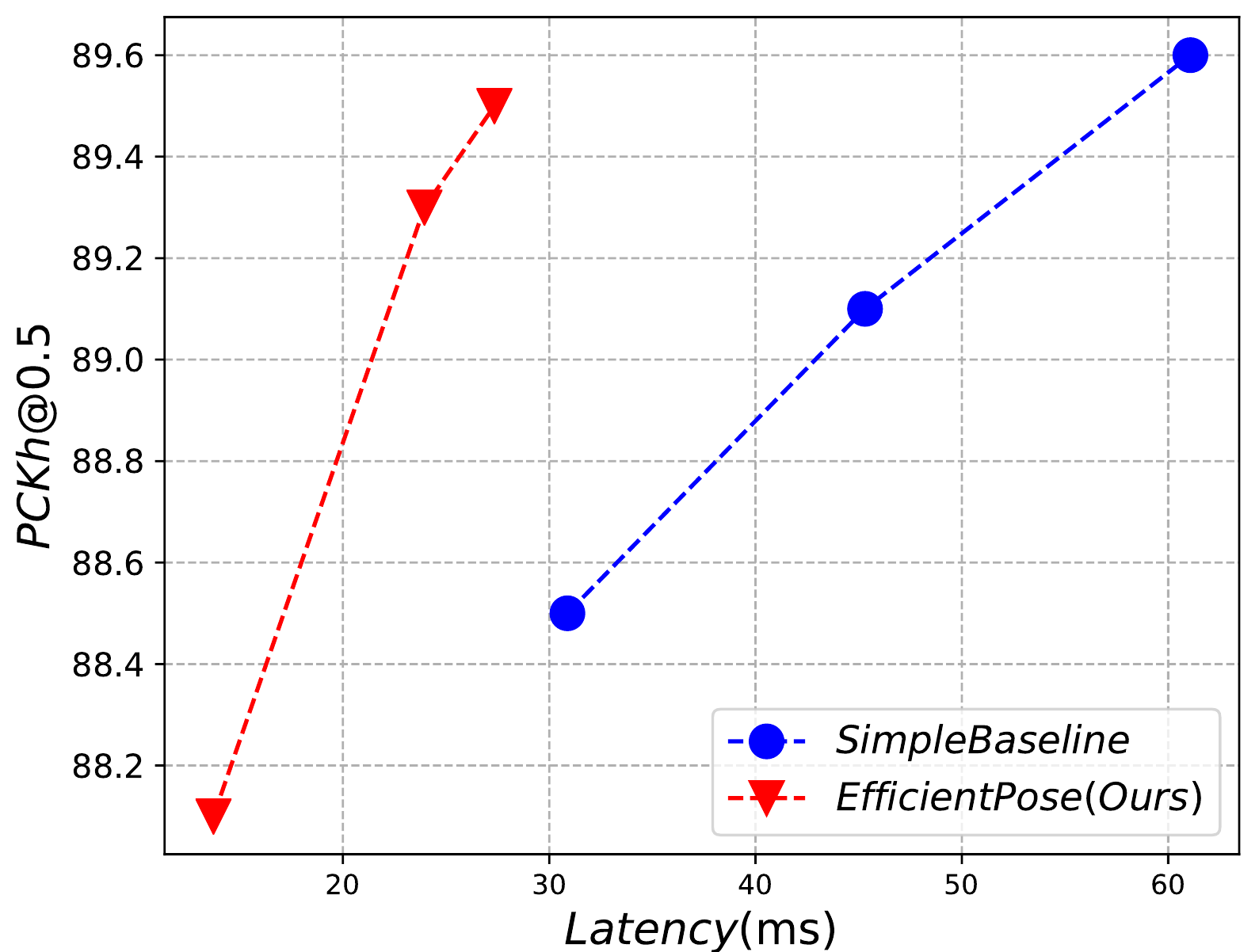}
\caption{GPU latency comparisons between EfficientPose and SimpleBaseline on the MPII val dataset. EfficientPose shows a far better trade-off between accuracy and latency.}
\label{fig: lat_comp}
\end{figure}

\begin{figure*}[t!]
    \centering
    \includegraphics[width=\linewidth]{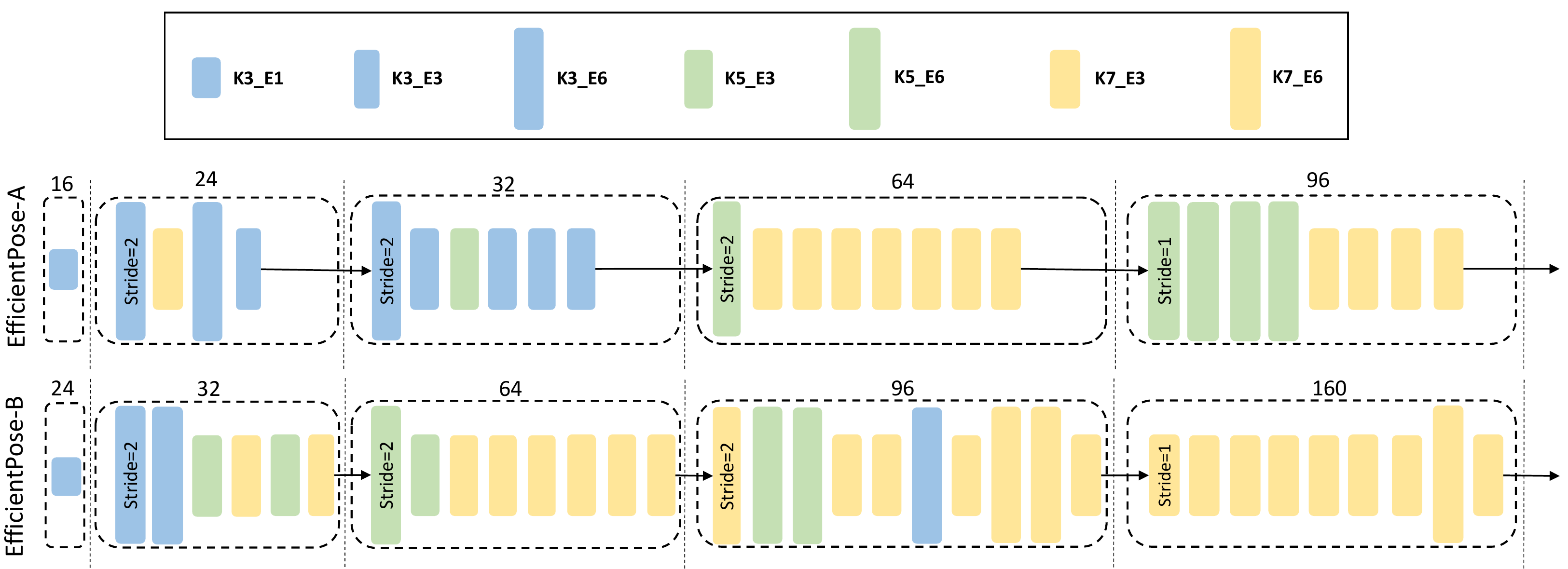}
    \caption{The architectures of EfficientPose-A and -B. We denote the MBConv blocks with diverse kernel sizes and expansion ratios as colored rectangle boxes. ``Kx\_Ey" denotes the block with kernel size ``x" and expansion ratio ``y". Blocks in the same stage (with the same width) are contained in the dashed box.}
    \label{fig: archs}
\end{figure*}

\paragraph{Comparisons with SOTA Pose Estimation Networks}
The results on the MPII validation dataset are shown in Tab.~\ref{tab: mpii results}. We search for three networks with different scales and compare them with the SOTA methods. Our EfficientPose networks achieve competitive and even higher performance with far less FLOPs. 
Compared with CPMs~\cite{wei2016cpm}, DLCM~\cite{tang2018deeply}, our EfficientPose-A takes only 0.7 GFLOPs and 13.73ms GPU latency while obtaining a similar accuracy. The cost of EfficientPose-B is only 1.5 GFLOPs and 23.95ms latency, while its performance surpasses other popular methods, e.g., Hourglass~\cite{newell2016stacked}, SimpleBaseline(ResNet101)~\cite{xiao2018simple}(45.31ms) and the knowledge distillation based method FPD~\cite{Zhang_2019_CVPR}. 
Compared with PNFS~\cite{yang2019pose}, which searches for the cell-based fabric architecture in the head part, EfficientPose-A and -B achieves higher or similar PCKh@0.5 with $2.9\times$ and $6.6\times$ fewer FLOPs. The performance of our large model EfficientPose-C is close to the latest SOTA networks, while its FLOPs is only 9.5\% of SimpleBaseline(ResNet152), 21.1\% of HRNet-W32~\cite{sun2019deep} and its latency is only 27.34ms, 44.8\% of SimpleBaseline(ResNet152) (61.08ms), 49.3\% of HRNet-W32~\cite{sun2019deep} (55.47ms). The GPU latency comparison is visualized in Fig.~\ref{fig: lat_comp}. And we show the searched architectures in Fig.~\ref{fig: archs}.

\begin{table}[thbp]
\centering
\caption{Comparisons with different backbone networks on the MPII validation set.}
    \small
    \begin{tabular}{lrrr}
    \toprule
    Backbone  & Params & GFLOPs & PCKh@0.5 \\
    \midrule
    MobileNetV2~\cite{sandler2018mobilenetv2}   & 1.5M & 0.73  & 86.63 \\
    Proxyless(GPU)~\cite{cai2018proxylessnas}   & 4.5M & 1.43  & 86.26 \\
    Proxyless(CPU)~\cite{cai2018proxylessnas}   & 1.8M & 1.05  & 87.54 \\
    Proxyless(Mobile)~\cite{cai2018proxylessnas} & 1.9M & 0.91  & 87.29 \\
    \midrule
    ResNet-50~\cite{he2016deep} & 25.6M & 9.70 & 88.02 \\
    \midrule
    Random Search     & 1.4M & 0.62  & 87.50 \\
    \midrule
    EfficientPose-A     & 1.3M    & 0.67  & 88.10 \\
    \bottomrule
    \end{tabular}
\label{tab: comp_backbone}
\vspace{-17pt}
\end{table}

\paragraph{Comparisons with Different Backbone Networks}
To further demonstrate the performance of our searched lightweight backbones, we compare our networks with others with only backbone networks changed, including manually engineered networks \cite{he2016deep,sandler2018mobilenetv2} and NAS networks~\cite{cai2018proxylessnas}. We train all the compared networks under the same training settings and hyperparameters. We only perform 4 down-sampling operations in the compared models for a fair comparison.

The comparison results are shown in Tab.~\ref{tab: comp_backbone}. In the first group, the models are all constructed with MBConv blocks \cite{sandler2018mobilenetv2}. EfficientPose-A achieves the highest PCKh@0.5 with the lowest FLOPs. Furthermore, compared with the large model ResNet-50~\cite{he2016deep}, EfficientPose-A achieves similar performance with 14.9$\times$ less FLOPs.

\begin{table}[thbp]
    \centering
    \caption{Generalization ability on the COCO validate set.}
    \resizebox{\columnwidth}{!}{
    \begin{tabular}{lcccc}
    \toprule
    Method      & Pretrain & Params & GFLOPs & AP \\
    \midrule
    Hourglass~\cite{newell2016stacked} & N & 25.1M       & 14.3   & 66.9 \\
    CPN~\cite{Chen2018CPN} & Y  & 102.0M    & 6.2    & 68.6 \\
    LPN-50~\cite{zhang2019simple}   & N  & 2.9M    & 1.0    & 68.9 \\ 
    EfficientPose-A     & N  & 1.3M  & 0.5  & 66.5 \\
    \midrule
    SimpleBaseline-R50~\cite{xiao2018simple}  & Y & 34.0M        & 8.9    & 70.4 \\
    LPN-101~\cite{zhang2019simple}    & N   & 5.3M       & 1.4    & 70.2 \\
    LPN-152~\cite{zhang2019simple}    & N   & 7.4M      & 1.8    & 70.8 \\
    EfficientPose-B     & N   & 3.3M   & 1.1     & 71.1 \\
    \midrule
    SimpleBaseline-R101~\cite{xiao2018simple} & Y  & 52.0M & 12.4   & 71.4 \\
    SimpleBaseline-R152~\cite{xiao2018simple} & Y  & 68.6M & 15.7   & 72.0 \\
    HRNet-W32~\cite{sun2019deep} & N  & 28.5M  & 7.1    & 73.4 \\
    MSPN~\cite{li2019rethinking}    & Y  & -   & 4.4    & 71.5  \\ 
    EfficientPose-C     & N  & 5.0M   & 1.6    & 71.3 \\
    \bottomrule
    \end{tabular}
    }
    \label{tab: coco_val}
\end{table}

\begin{table*}[thbp]
    \centering
    \caption{Generalization ability on the COCO test set.}
    \resizebox{\textwidth}{!}{
    \begin{tabular}{lllrcrrrrrr}
    \toprule
    Method  & Backbone & Input size & Params & GFLOPs & AP & AP$^{50}$ & AP$^{75}$  & AP$^{M}$  & AP$^{L}$  & AR\\
    \midrule
    Mask-RCNN~\cite{he2017mask}  & ResNet-50-FPN  & -  & -  & -  & 63.1  & 87.3  & 68.7  & 57.8  & 71.4  & - \\
    G-RMI~\cite{pap2017accurate}  & ResNet-101  & 353 $\times$ 257  & 42.6M  & 57.0  & 64.9  & 85.5  & 71.3  & 62.3  & 70.0  & 69.7 \\
    G-RMI + extra data~\cite{pap2017accurate}  & ResNet-101  & 353 $\times$ 257  & 42.6M  & 57.0  & 68.5  & 87.1  & 75.5  & 65.8  & 73.3  & 73.3 \\ 
    CFN~\cite{huang2017coarse}  & -  & 384 $\times$ 288  & -  & -  & 72.6  & 86.1  & 69.7  & 78.3  & 64.1  & -  \\
    RMPE~\cite{yang2017pyramid}  & PyraNet  & 320 $\times$ 256  & 28.1M  & 26.7  & 72.3  & 89.2  & 79.1  & 68.0  & 78.6  & - \\
    CPN~\cite{Chen2018CPN}  & ResNet-Inception  & 384 $\times$ 288  & -  & -  & 72.1  & 91.4  & 80.0  & 68.7  & 77.2  & 78.5 \\
    SimpleBaseline~\cite{xiao2018simple}  & ResNet-50  & 256 $\times$ 192  & 34.0M  & 8.9  & 70.0  & 90.9 & 77.9  & 66.8  & 75.8  & 75.6 \\
    SimpleBaseline~\cite{xiao2018simple}  & ResNet-152  & 256 $\times$ 192  & 68.6M  & 15.7  & 71.6  & 91.2  & 80.1  & 68.7  & 77.2  & 77.3 \\
    HRNet-W32~\cite{sun2019deep}  & HRNet-W32  & 384 $\times$ 288  & 28.5M  & 16.0  & 74.9  & 92.5  & 82.8  & 71.3  & 80.9  & 80.1 \\
    HRNet-W48~\cite{sun2019deep}  & HRNet-W48  & 384 $\times$ 288  & 63.6M  & 32.9  & 75.5  & 92.5  & 83.3  & 71.9  & 81.5  & 80.5 \\ 
    LPN~\cite{zhang2019simple}  & ResNet-50  & 256 $\times$ 192  & 2.9M  & 1.0  & 68.7  & 90.2  & 76.9  & 65.9  & 74.3  & 74.5 \\
    LPN~\cite{zhang2019simple}  & ResNet-101  & 256 $\times$ 192  & 5.3M  & 1.4  & 70.0  & 90.8  & 78.4  & 67.2  & 75.4  & 75.7 \\
    LPN~\cite{zhang2019simple}  & ResNet-152  & 256 $\times$ 192  & 7.4M  & 1.8  & 70.4  & 91.0  & 78.9  & 67.7  & 76.0  & 76.2  \\
    PNFS~\cite{yang2019pose} & MobileNet-V2 & 256 $\times$ 192 & 6.1M & 4.0 & 67.4 & 89.0 & 73.7 & 63.3 & 74.3 & 73.1 \\
    PNFS~\cite{yang2019pose} & ResNet-50 & 256 $\times$ 192 & 27.5M & 11.4 & 70.9 & 90.4 & 77.7 & 66.7 & 78.2 & 76.6 \\
    \hline
    EfficientPose-B    & NAS searched  & 256 $\times$ 192    & 3.3M  & 1.1    & 70.5    & 91.1    & 79.0    & 67.3    & 76.2    & 76.1 \\
    EfficientPose-C    & NAS searched  & 256 $\times$ 192    & 5.0M  & 1.6    & 70.9    & 91.3    & 79.4    & 67.7    & 76.5    & 76.5 \\
    \bottomrule
    \end{tabular}
    }
    \label{tab: coco_test}
    \vspace{-7pt}
\end{table*}

\subsection{Generalization Ability on COCO}
We implement our models searched on MPII to COCO~\cite{COCO}. Only the output dimension of the final 1$\times$1 convolution is adjusted as the number of keypoints changes. COCO contains about 200K images with about 250K person instances. We adopt the same data augmentation strategy as HRNet~\cite{sun2019deep}. The input size is set as $192\times256$.

The whole training process takes 240 epochs using the Adam optimizer with the batch size of 128. We use the warm-up strategy in the first 500 iterations to linearly increase the learning rate to 1e-3. Then the learning rate is decayed by 0.1 at 200K and 240K iteration respectively. 

\begin{figure*}[thbp]
    \centering
    \includegraphics[width=\textwidth]{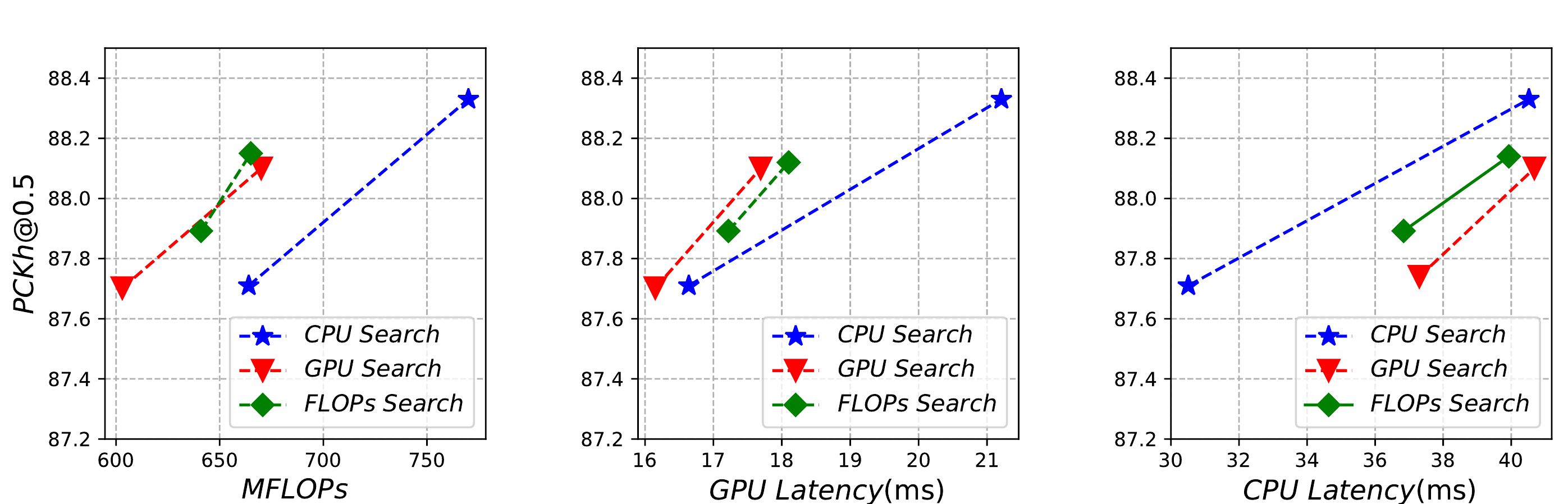}
    \caption{Specialization results of different model cost benchmarks on the MPII validation set. The networks optimized with different benchmarks are denoted with different colors.}
    \label{fig: hardwares}
\end{figure*}

As shown in Tab.~\ref{tab: coco_val}, our EfficientPose networks achieve high AP with lowest FLOPs on COCO. With nearly 500 MFLOPs, the EfficientPose-A network achieves a comparable result compared with Hourglass~\cite{newell2016stacked}, CPN~\cite{Chen2018CPN} and LPN~\cite{zhang2019simple}. EfficientPose-B surpasses both LPN~\cite{zhang2019simple} networks and SimpleBaseline-R50 with only 1.1 GFLOPs. EfficientPose-C also obtains a high AP with far less FLOPs. The effectiveness of EfficientPose networks is demonstrated on the COCO test set in Tab.~\ref{tab: coco_test} as well.

\subsection{Specialization on Different Hardwares}
We specialize our EfficientPose network design on three different model cost benchmarks, i.e., FLOPs, GPU latency and CPU latency. We achieve this by remeasure the cost term in the loss function (Eq.~\ref{eq: loss}) of the backbone architecture search. The GPU latency is measured on one Tesla V100 GPU with batch size 32 and the CPU latency is measured on 1 Intel(R) Core(TM) i7-8700K CPU with a batch size of 1. For each cost benchmark, we search for two networks and visualize the results in Fig.~\ref{fig: hardwares}. This experiment shows that our pose estimation framework can be efficient on diverse hardware platforms. The specialization process can be easily performed by building different lookup tables on the cost benchmarks for predicting the model cost. Though FLOPs is widely used for evaluating the cost of the model, the results in Fig.~\ref{fig: hardwares} and some recent works~\cite{MnasNet,cai2018proxylessnas} demonstrate that FLOPs is not so correlated with the real inference speed due to the discrepancy of different hardware platforms. The ability to specialize models on the target device is so vital for real applications.

\subsection{Ablation Studies}
\begin{figure}[htbp]
    \centering
    \subfigure[GT]{
    \begin{minipage}[b]{0.23\linewidth}
        \includegraphics[width=\linewidth]{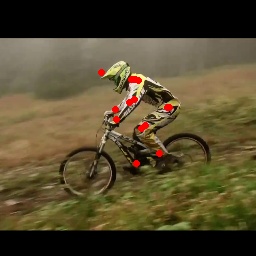}\vspace{4pt}
        \includegraphics[width=\linewidth]{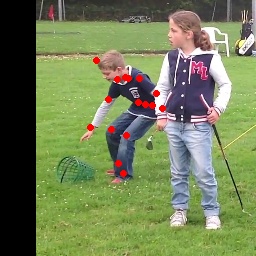}
    \end{minipage}}
    \subfigure[w/o SIC]{
    \begin{minipage}[b]{0.23\linewidth}
        \includegraphics[width=\linewidth]{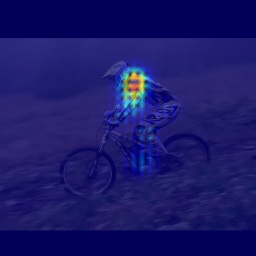}\vspace{4pt}
        \includegraphics[width=\linewidth]{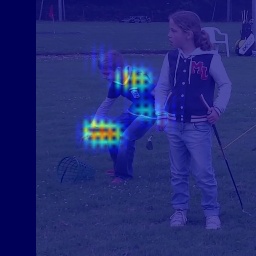}
    \end{minipage}}
    \subfigure[before SIC]{
    \begin{minipage}[b]{0.23\linewidth}
        \includegraphics[width=\linewidth]{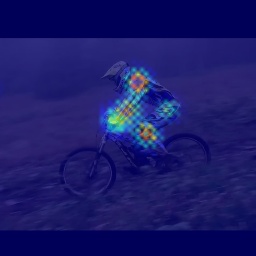}\vspace{4pt}
        \includegraphics[width=\linewidth]{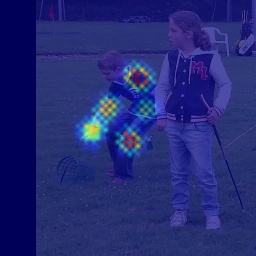}
    \end{minipage}}
    \subfigure[after SIC]{
    \begin{minipage}[b]{0.23\linewidth}
        \includegraphics[width=\linewidth]{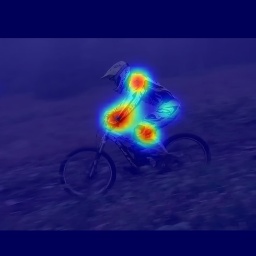}\vspace{4pt}
        \includegraphics[width=\linewidth]{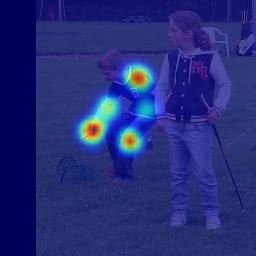}
    \end{minipage}}
    \caption{Visualization of feature maps in the network. (a) the ground truth. (b) the feature maps after the transposed convolutions in the network trained without SIC. (c) the feature maps before SIC in the EfficientPose network. (d) the feature maps after SIC in the EfficientPose network.}
    \label{fig: vis_feature}
\end{figure}

\paragraph{Effectiveness of SIC module}
We visualize the feature maps in the network to study the effect of the spatial information correction (SIC) module in Fig.~\ref{fig: vis_feature}. We find that there exists the checkerboard pattern of artifacts~\cite{odena2016deconvolution} in both the feature maps after the transposed convolutions in the network trained without SIC and the feature maps before SIC in EfficientPose networks. The SIC module eliminates the checkerboard pattern in the feature maps obtained by transposed convolutions and makes the interested field more concentrated, which contributes to the final prediction. We further perform the ablation studies of SIC in Tab.~\ref{tab: sic}. The SIC module shows evident accuracy promotion in both the manually designed network MobileNetV2 and the EfficientPose networks.

\begin{table}[thbp]
    \centering
    \caption{Ablation Studies of the SIC module.}
    \small
    \begin{tabular}{ccl}  
    \toprule
    Model   & SIC    & PCKh@0.5 \\
    \midrule
    \multirow{2}{*}{MobileNetv2~\cite{sandler2018mobilenetv2}} & & 86.34 \\
    & \checkmark & 86.63$_{\uparrow 0.29}$ \\
                \hline
    \multirow{2}{*}{EfficientPose-A}   &     & 87.62 \\
                &  \checkmark    & 88.10$_{\uparrow 0.48}$  \\
                \hline
    \multirow{2}{*}{EfficientPose-B}  &    & 88.55 \\
                &  \checkmark   & 89.27$_{\uparrow 0.72}$   \\
                \hline
    \multirow{2}{*}{EfficientPose-C}  &    & 88.61  \\   
                & \checkmark    & 89.49$_{\uparrow 0.88}$   \\
    \bottomrule
    \end{tabular}
    \label{tab: sic}
    \vspace{-7pt}
\end{table}

\begin{table}[thbp]
\caption{Comparisons with larger width settings of the transposed convolutions. TConv denotes the transposed convolution.}
\centering
\small
    \begin{tabular}{ccccl}
    \toprule
    \multicolumn{2}{c}{Width} & \multicolumn{2}{c}{MFLOPs} & \multirow{2}*{PCKh@0.5} \\
    TConv1 & TConv2 & Backbone & Head & \\
    \midrule
    64 & 32 & \multirow{3}*{405} & 264  & 88.100\\
    64 & 64 &  & 369  & 88.095  \\
    96 & 96 &  & 755  & 87.918  \\
    \bottomrule
    \end{tabular}
\label{tab:head_width}
\vspace{-10pt}
\end{table}

\paragraph{Efficiency of Slim Transposed Convolutions}
We set the widths of the two transposed convolutions in the head as 64 and 32 respectively, which are always set larger in previous works, e.g., 256 in SimpleBaseline~\cite{xiao2018simple}. To further demonstrate the efficiency of our slim transposed convolutions, we compare with larger width settings of the two transpose convolutions in the head. As shown in Tab.~\ref{tab:head_width}, when we enlarge the widths to 64 and 64, the FLOPs in the head increase 134M but no accuracy promotion is obtained. When we set the widths larger, 96 and 96, the FLOPs increase 520M and a worse PCKh@0.5 is get. It is worth noting the FLOPs of the (96, 96) head is much larger than the backbone. We attribute the performance degradation to overfitting when the head is too heavy with a lightweight backbone.

\begin{table}[thbp]
\caption{Studies of more efficient transposed convolutions.}
\centering
\small
\begin{tabular}{ccl}  
\toprule
Transposed Conv  & MFLOPs    & PCKh@0.5 \\
\midrule
Plain    & 669  & 88.10  \\
MBV1 style   & 469  & 87.91  \\
MBV2 style   & 600  & 87.94   \\
\bottomrule
\end{tabular}
\label{tab: trans_conv}
\vspace{-16pt}
\end{table}    

\paragraph{Studies on Efficient Transposed Convolutions}
We study lighter convolution modules for the transposed convolutions, the separable depthwise convolution in MobileNetV1~\cite{howard2017mobilenets} (MBV1) and the inverted residual block in MobileNetV2~\cite{sandler2018mobilenetv2} (MBV2). As shown in Tab.~\ref{tab: trans_conv}, we find that the MBV1 style module decreases the FLOPs by 181M with acceptable accuracy decay, which can be an option for the efficient head as well.

\paragraph{Comparisons with Random Search}
We perform the random search experiment, which is a vital baseline for evaluating the effectiveness of the NAS method~\cite{li2019random}. We randomly sample 50 architectures and train each one for 5 epochs to evaluate the validation performance. Finally, we select the best-performance one and train it with the same settings as our EfficientPose. The total cost of random search is the same as ours. The results are shown in Tab.~\ref{tab: comp_backbone}. Our EfficientPose-A shows an evident advantage over the random searched one.

\begin{figure*}[htbp]
    \centering
    \subfigure{
    \begin{minipage}[b]{0.3\linewidth}
        \includegraphics[width=1\linewidth]{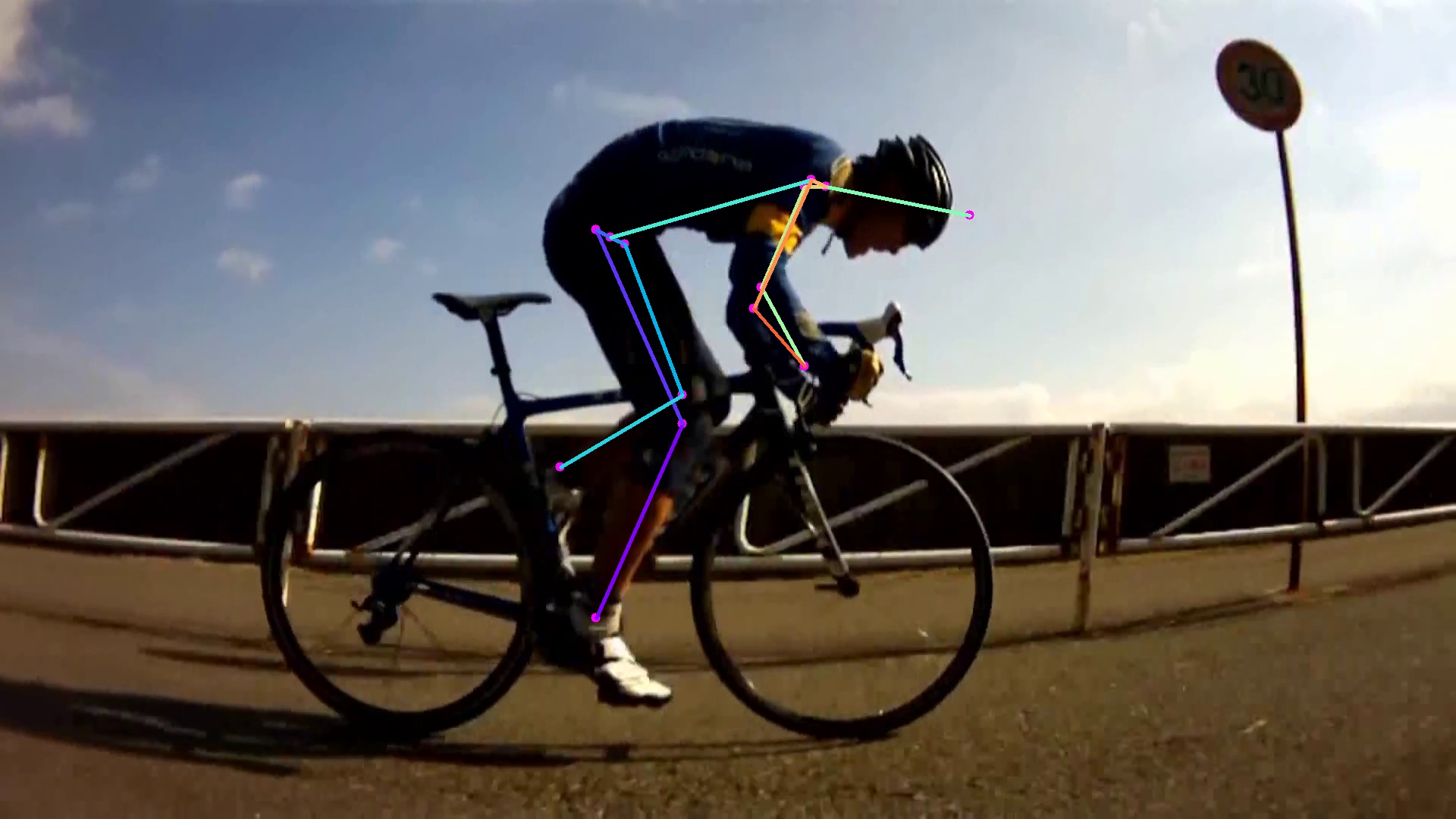}\vspace{4pt}
        \includegraphics[width=1\linewidth]{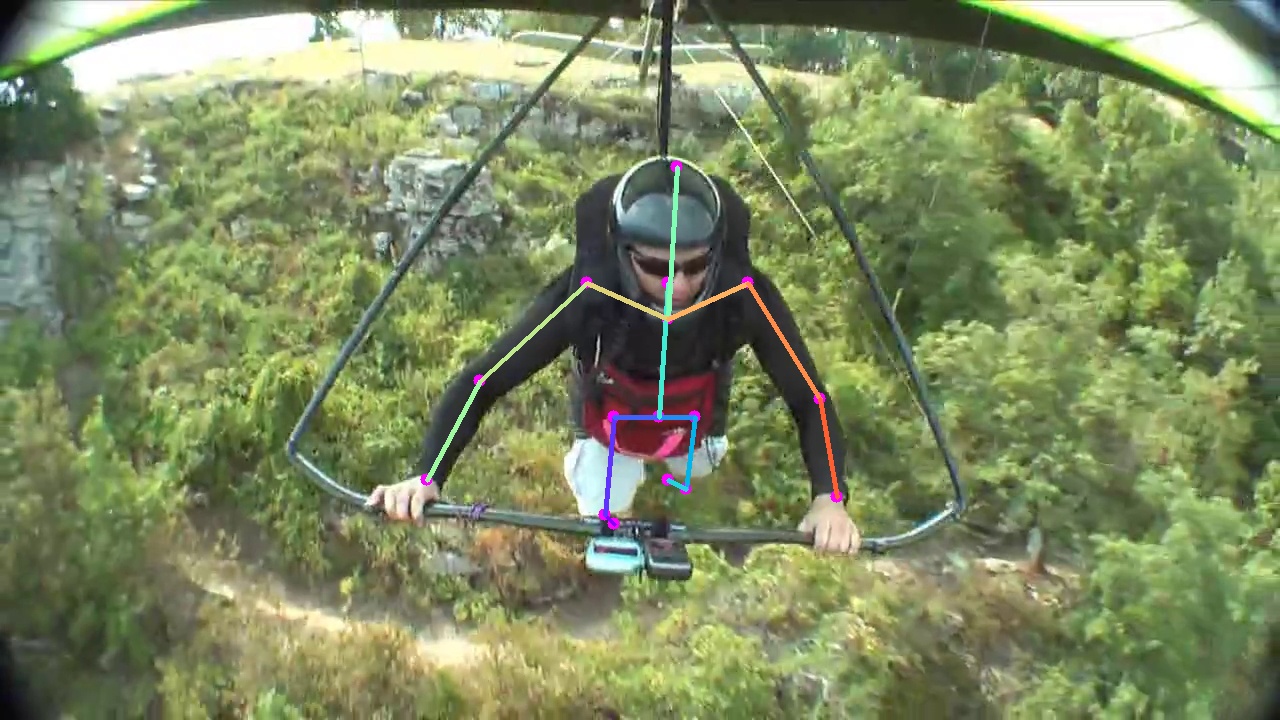}
    \end{minipage}}
    \subfigure{
    \begin{minipage}[b]{0.3\linewidth}
        \includegraphics[width=1\linewidth]{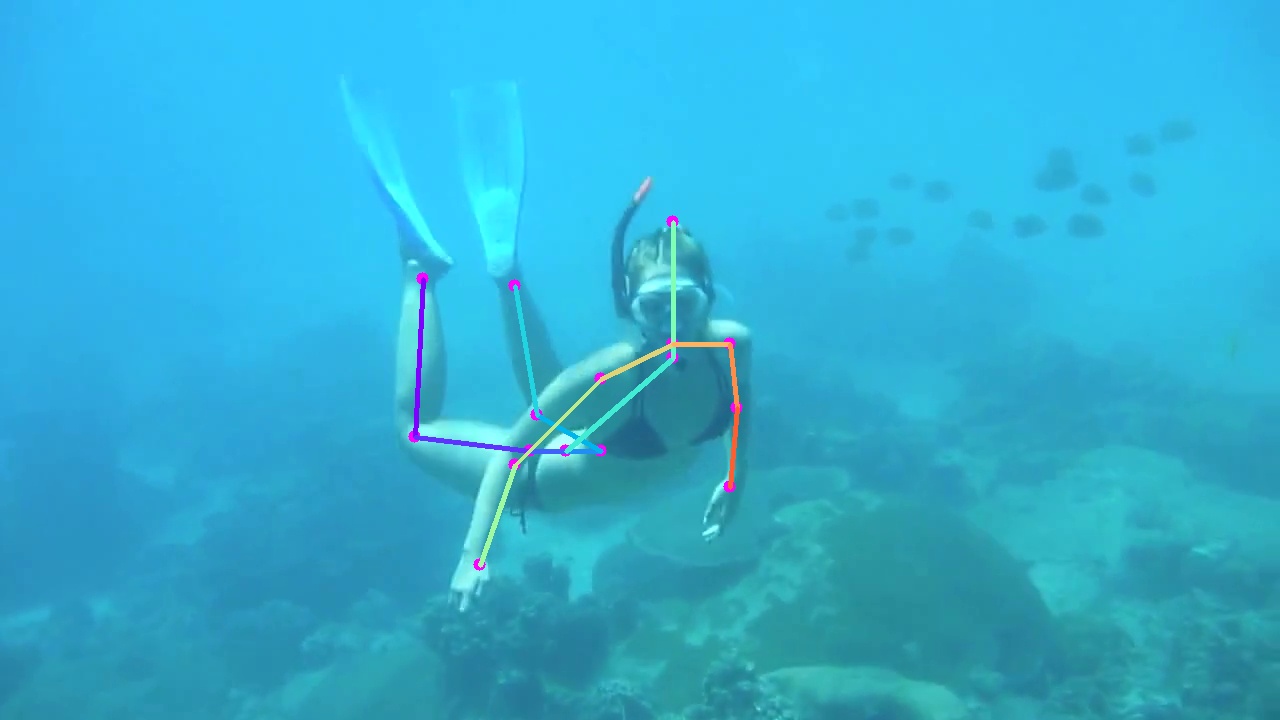}\vspace{4pt}
        \includegraphics[width=1\linewidth]{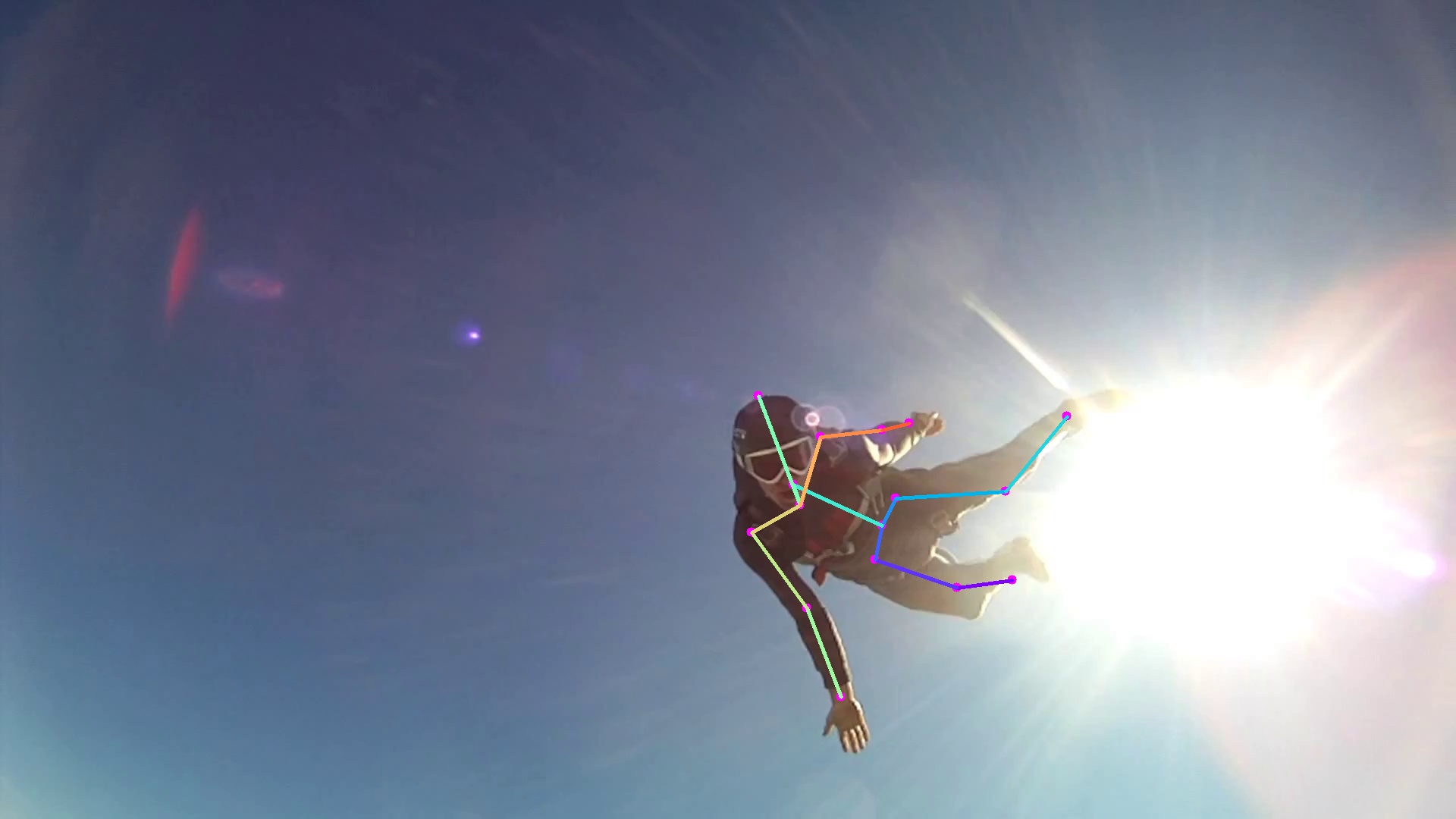}
    \end{minipage}}
    \subfigure{
    \begin{minipage}[b]{0.3\linewidth}
        \includegraphics[width=1\linewidth]{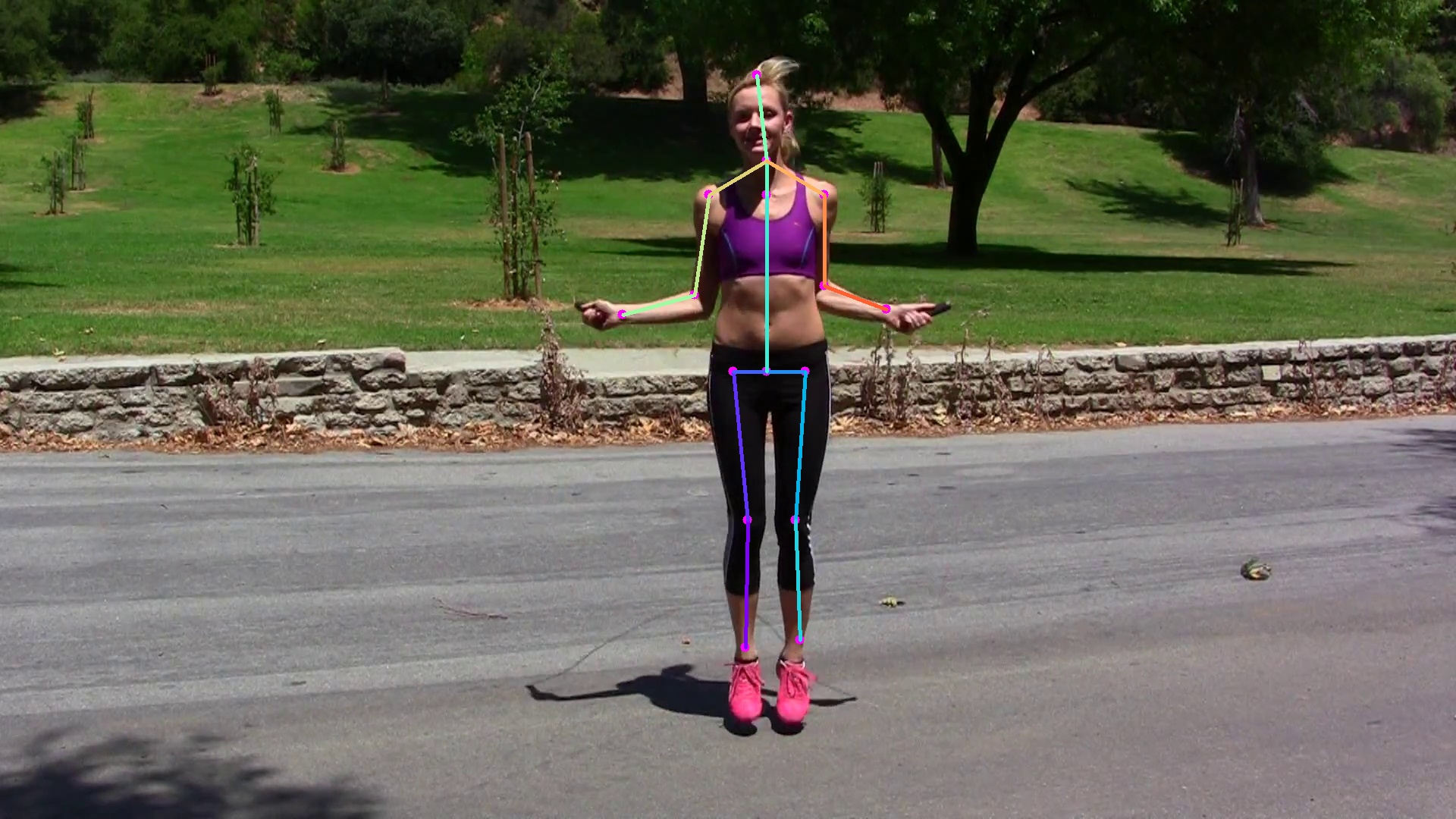}\vspace{4pt}
        \includegraphics[width=1\linewidth]{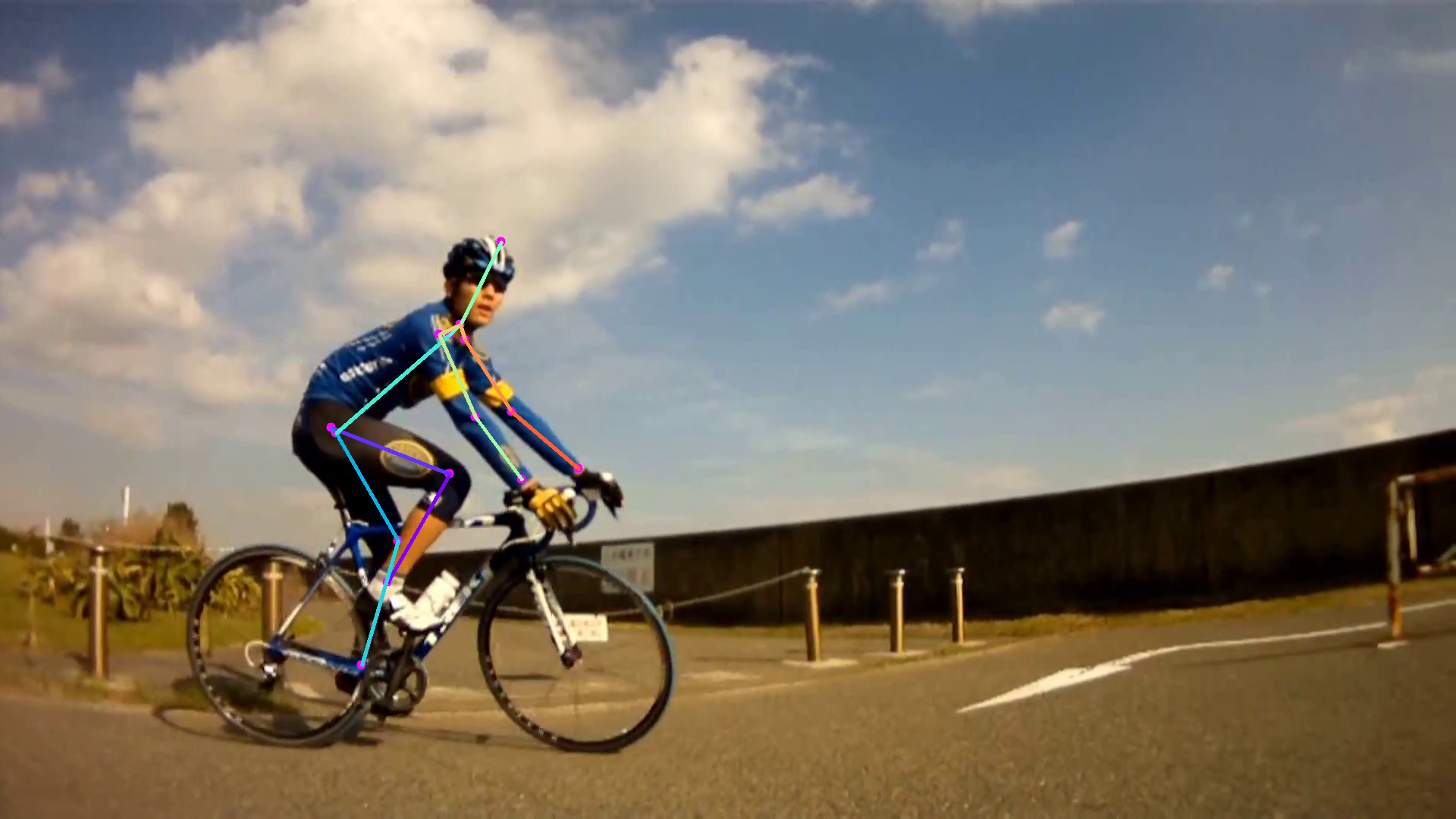}
    \end{minipage}}
    \caption{Visualization results on the MPII validation set.}
    \label{fig: vis_mpii}
\end{figure*}

\begin{figure*}[htbp]
    \centering
    \subfigure{
    \begin{minipage}[b]{0.3\linewidth}
        \includegraphics[width=1.0\linewidth]{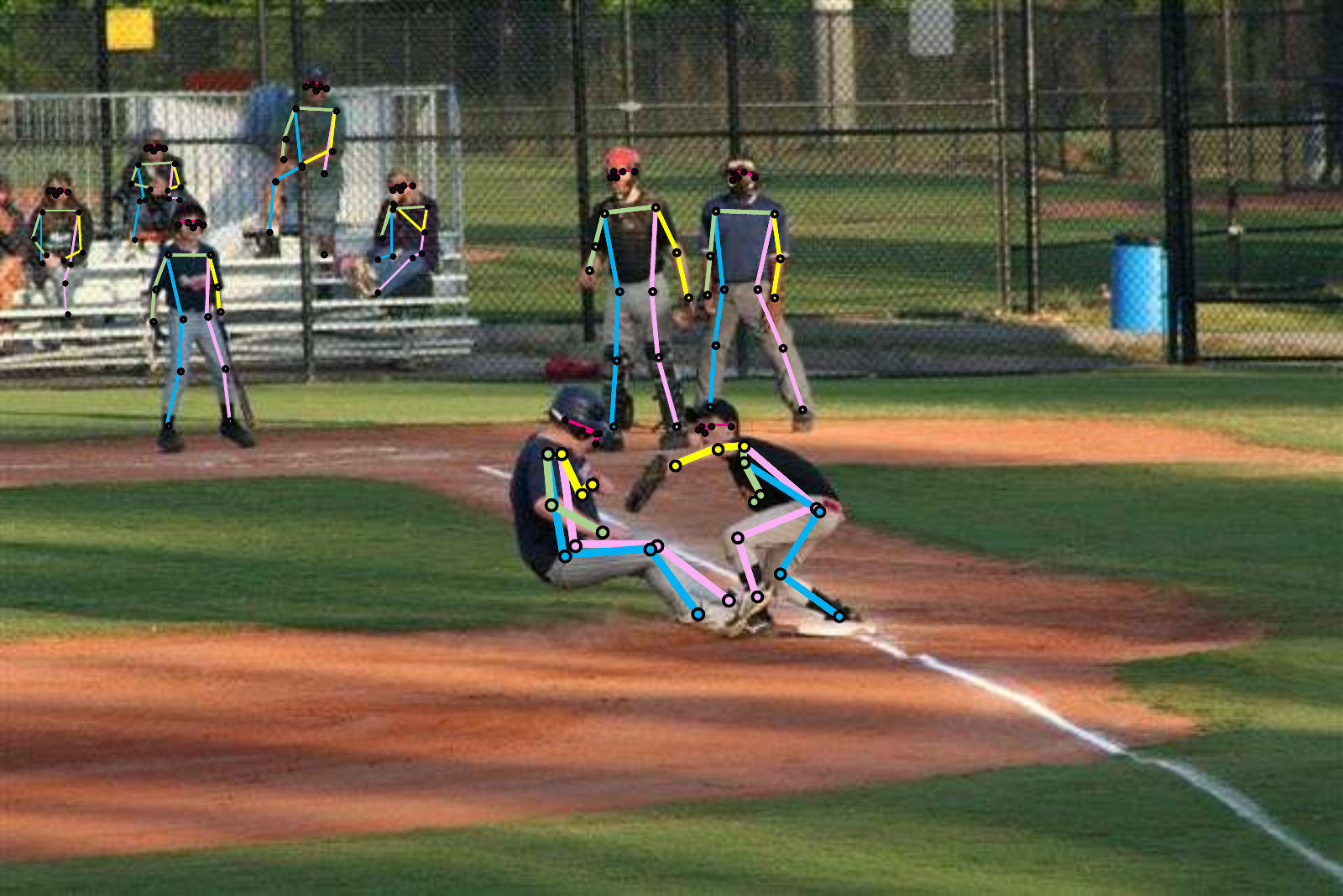}\vspace{4pt}
        \includegraphics[width=1\linewidth]{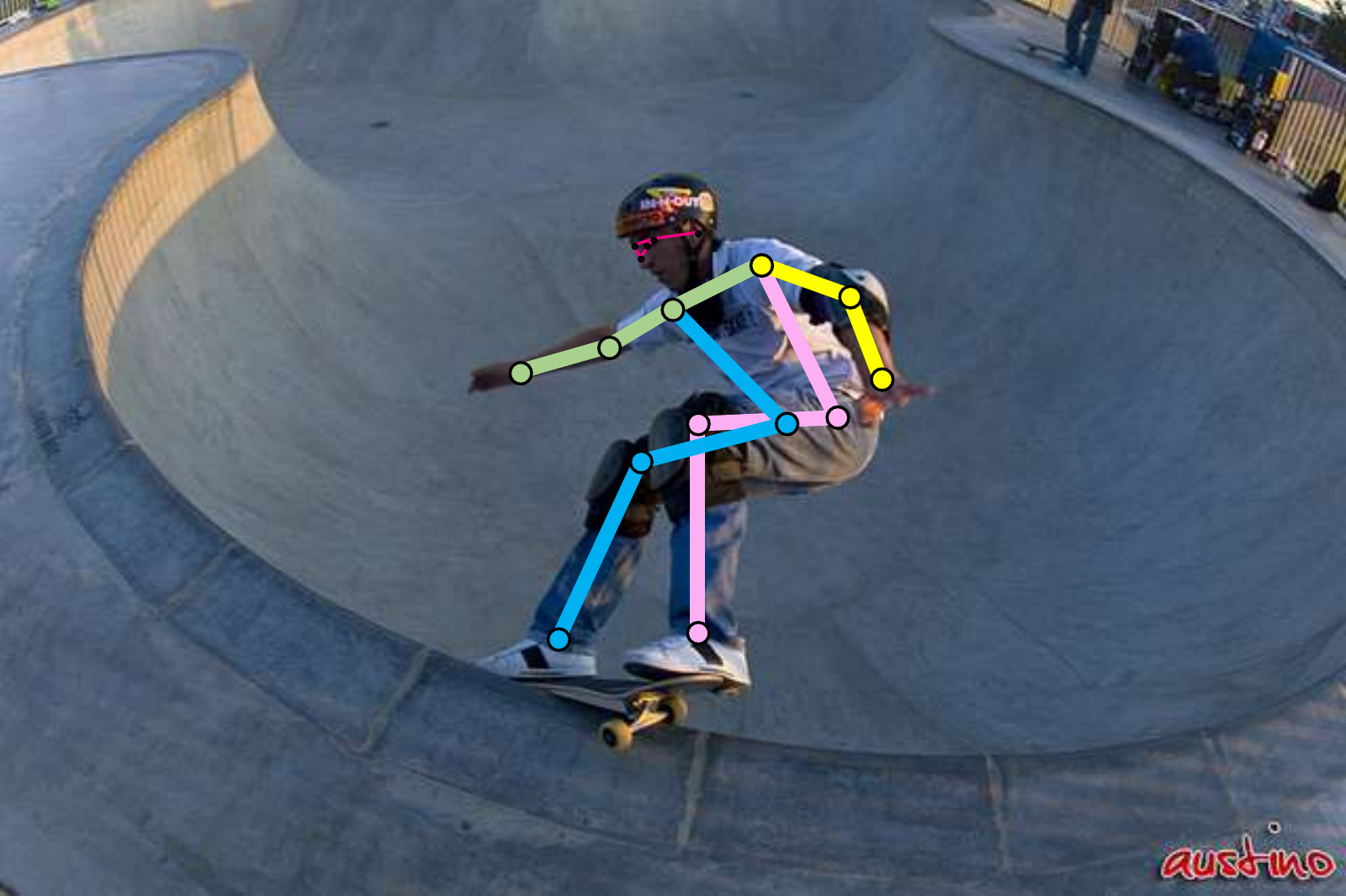}
    \end{minipage}}
    \subfigure{
    \begin{minipage}[b]{0.3\linewidth}
        \includegraphics[width=0.935\linewidth]{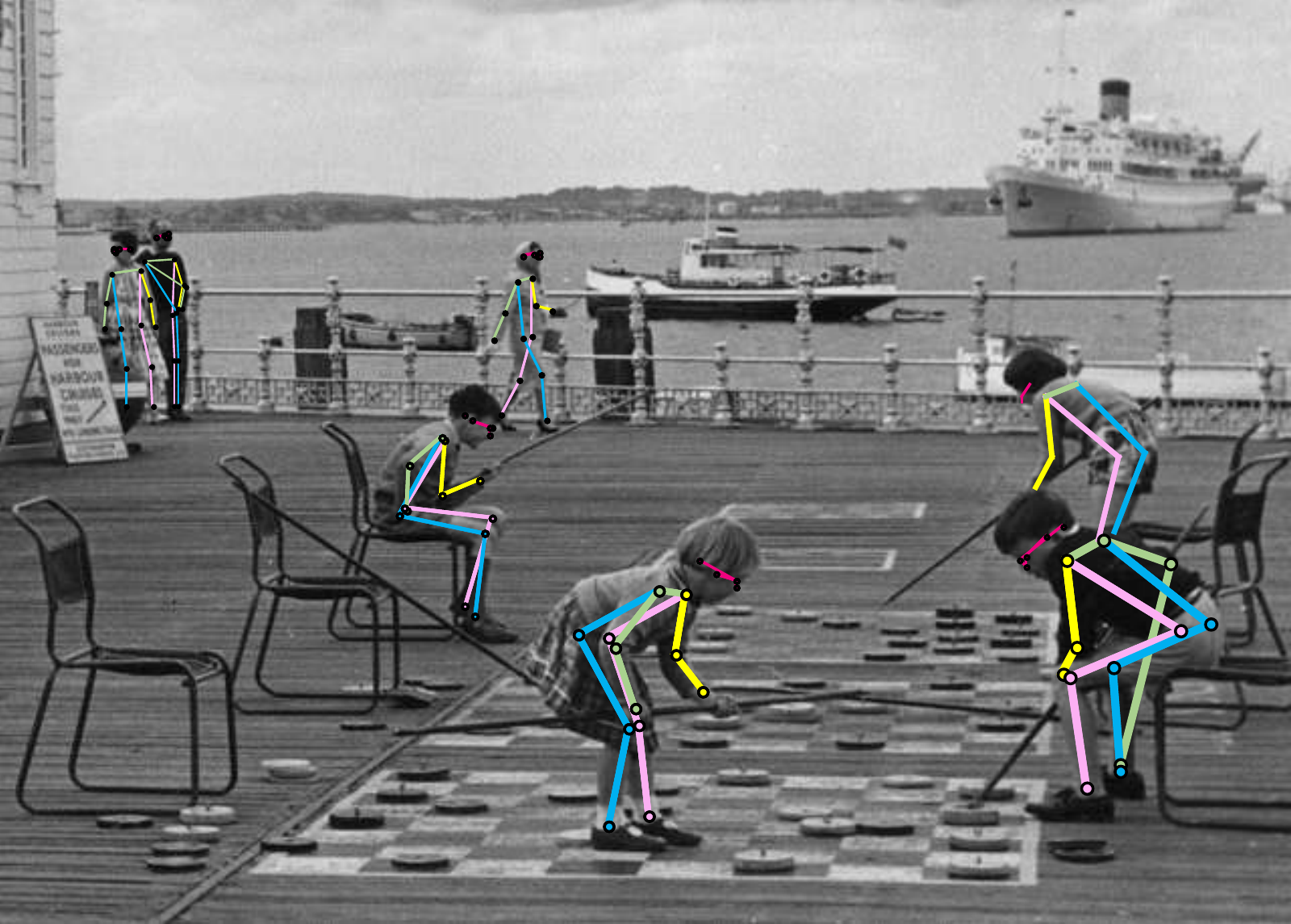}\vspace{4pt}
        \includegraphics[width=1\linewidth]{imgs/coco/id_80659_4_compressed.pdf}
    \end{minipage}}
    \subfigure{
    \begin{minipage}[b]{0.3\linewidth}
        \includegraphics[width=1.\linewidth]{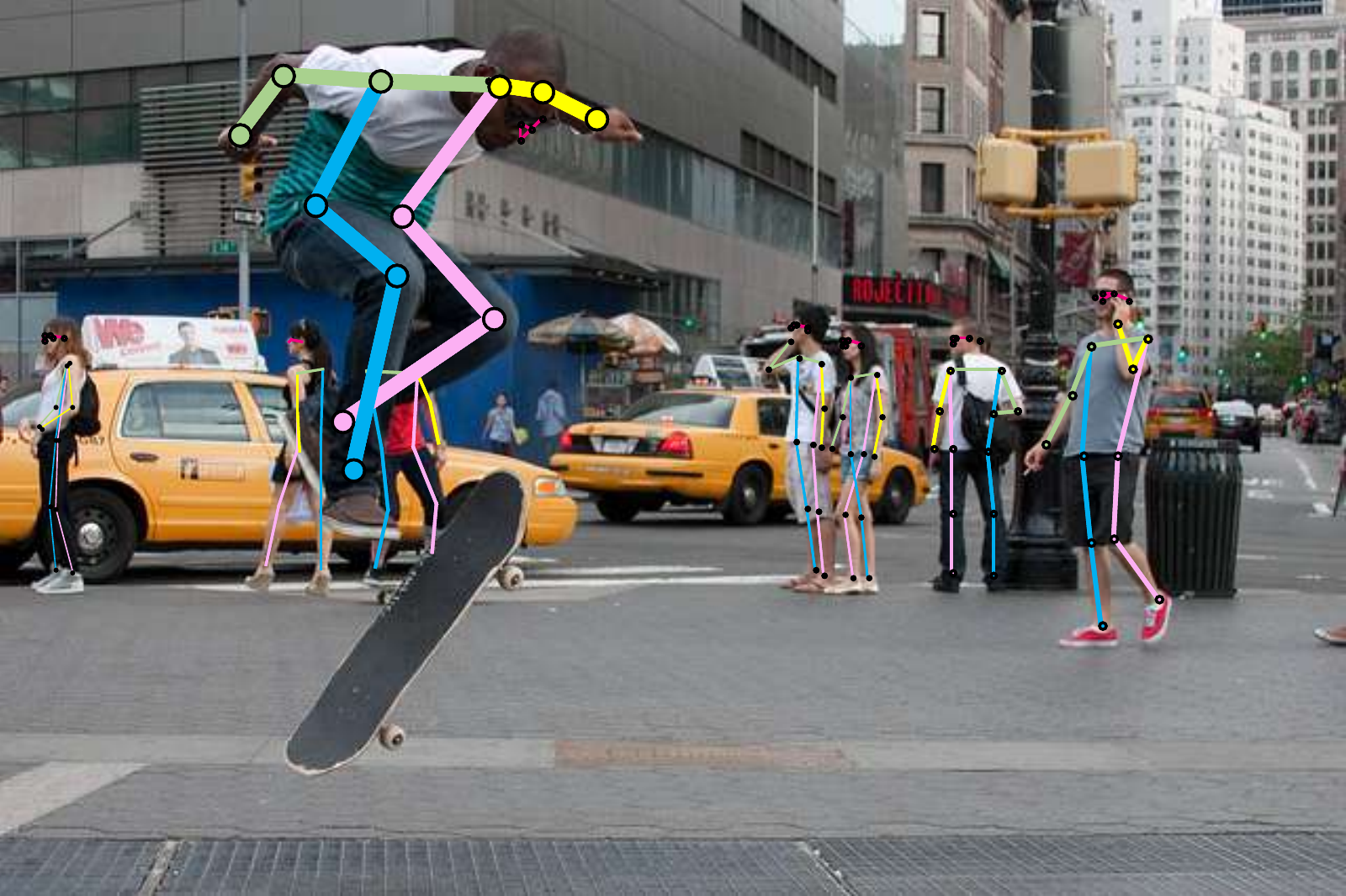}\vspace{4pt}
        \includegraphics[width=1\linewidth]{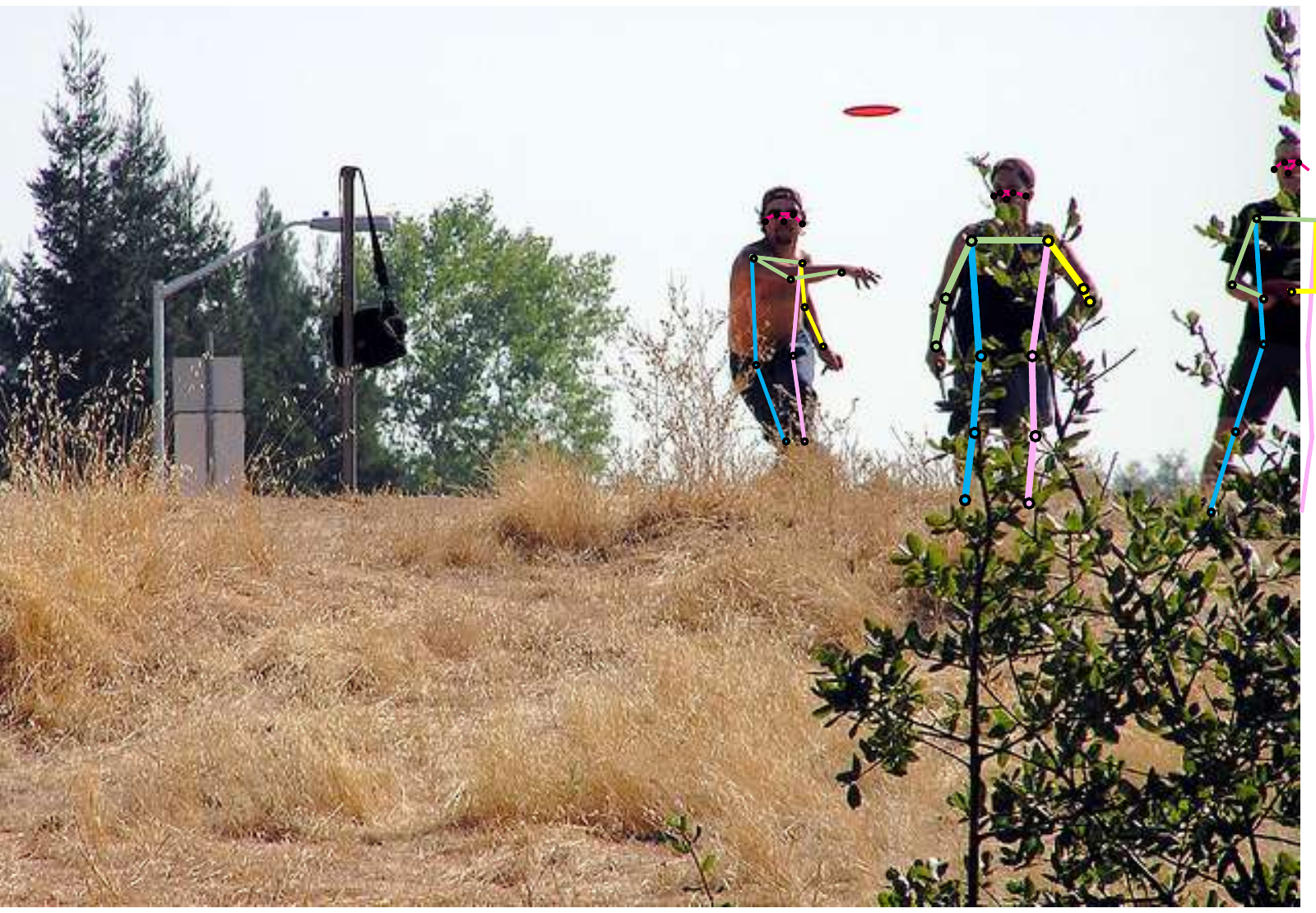}
    \end{minipage}}
    \caption{Visualization results on the COCO validation set.}
    \label{fig: vis_coco}
    \vspace{-7pt}
\end{figure*}

\section{Conclusion}
In this paper, we propose a framework targeted at efficient human pose estimation. We use the differentiable NAS method to automatically customize the backbone network for pose estimation and reduce the computation cost greatly. We further design an efficient head network which includes both the slim transposed convolutions and a spatial information correction module to promote the prediction performance with negligible FLOPs/latency increase. The proposed EfficientPose networks achieve similar accuracies with far less computation cost compared to other SOTA methods.

{\small
\bibliographystyle{ieee_fullname}
\bibliography{references}
}

\end{document}